\documentclass[letterpaper]{article} 
\usepackage{aaai25}  
\usepackage{times}  
\usepackage{helvet}  
\usepackage{courier}  
\usepackage[hyphens]{url}  
\usepackage{graphicx} 
\usepackage{natbib}  
\usepackage{caption} 
\usepackage{graphicx}
\usepackage{booktabs}
\usepackage{multirow}
\usepackage{bbding}
\usepackage{pifont}
\usepackage{amsmath}
\usepackage{tablefootnote}
\usepackage{amssymb}
\usepackage{appendix}
\usepackage[ruled,vlined]{algorithm2e}

\newcommand{\pluseq}{\mathrel{+}=}
\newcommand{\minuseq}{\mathrel{-}=}

\usepackage{newfloat}
\usepackage{listings}
\DeclareCaptionStyle{ruled}{labelfont=normalfont,labelsep=colon,strut=off} 
\title{VQTalker: Towards Multilingual Talking Avatars \\ through Facial Motion Tokenization}

\author{
    Tao Liu\textsuperscript{\rm 1}\thanks{Completed during the author's internship.}, Ziyang Ma\textsuperscript{\rm 1}, Qi Chen\textsuperscript{\rm 1}, Feilong Chen\textsuperscript{\rm 2}, Shuai Fan\textsuperscript{\rm 2}, Xie Chen\textsuperscript{\rm 1}, Kai Yu\textsuperscript{\rm 1}\thanks{The Corresponding author.}
}
\affiliations{
    \textsuperscript{\rm 1}X-LANCE Lab, MoE Key Lab of Artificial Intelligence, Shanghai Jiao Tong University
    \textsuperscript{\rm 2}AISpeech Ltd
}

\usepackage{bibentry}

\begin{document}

\maketitle
\begin{abstract}

We present VQTalker, a \textbf{V}ector \textbf{Q}uantization-based framework for multilingual talking head generation that addresses the challenges of lip synchronization and natural motion across diverse languages. Our approach is grounded in the phonetic principle that human speech comprises a finite set of distinct sound units (phonemes) and corresponding visual articulations (visemes), which often share commonalities across languages. We introduce a facial motion tokenizer based on Group Residual Finite Scalar Quantization (GRFSQ), which creates a discretized representation of facial features. This method enables comprehensive capture of facial movements while improving generalization to multiple languages, even with limited training data. Building on this quantized representation, we implement a coarse-to-fine motion generation process that progressively refines facial animations. Extensive experiments demonstrate that VQTalker achieves state-of-the-art performance in both video-driven and speech-driven scenarios, particularly in multilingual settings. Notably, our method achieves high-quality results at a resolution of $512 \times 512$  pixels while maintaining a lower bitrate of approximately 11 kbps. Our work opens new possibilities for cross-lingual talking face generation.

\end{abstract}

\section{Introduction}

Audio-driven talking head generation aims to create realistic facial animation synchronized with input audio and has various applications in film dubbing and animation production. The core challenge lies in precise lip synchronization, which is crucial due to the McGurk effect~\cite{mcgurk1976hearing}. This effect demonstrates the intricate interaction between auditory and visual cues in human speech perception. Although current audio-driven talking head generation methods have made significant progress, existing algorithms~\cite{sadtalker, ma2023styletalk, anitalker} still face synchronization issues, such as misalignment between audio and visual cues or the absence of specific lip shapes. These problems are particularly pronounced when dealing with languages outside the Indo-European family.

This limitation primarily stems from two factors. Commonly used training datasets like VoxCeleb \cite{voxceleb}, CelebV-HQ~\cite{zhu2022celebv}, MEAD~\cite{mead}, and HDTF \cite{hdtf} are overwhelmingly dominated by Indo-European languages, like English, German, and French. As a result, models trained on these datasets perform well within the language family but often fail to accurately capture the lip movements and facial expressions associated with other linguistic groups.
Secondly, the reliance on continuous representations in existing methods further compounds the problem. Those approaches typically operate in continuous spaces, such as the continuous spectral representations of audio inputs and the continuous latent embeddings of facial motion outputs. However, different language families possess a finite number of distinct phonetic structures and visual articulations. Continuous representations allow for infinite variations, potentially increasing training difficulty and leading to overfitting specific languages.

While some approaches \cite{yang2020large, huang2021speaker, song2022talking, sung2024multitalk} focus on increasing training data volume for other languages, this strategy is resource-intensive and particularly challenging for minority languages. Moreover, simply increasing data volume may not fully address the underlying need for models that can generalize across diverse linguistic systems. 

\begin{figure}[t]
    \centering
    \includegraphics[width=1\linewidth]{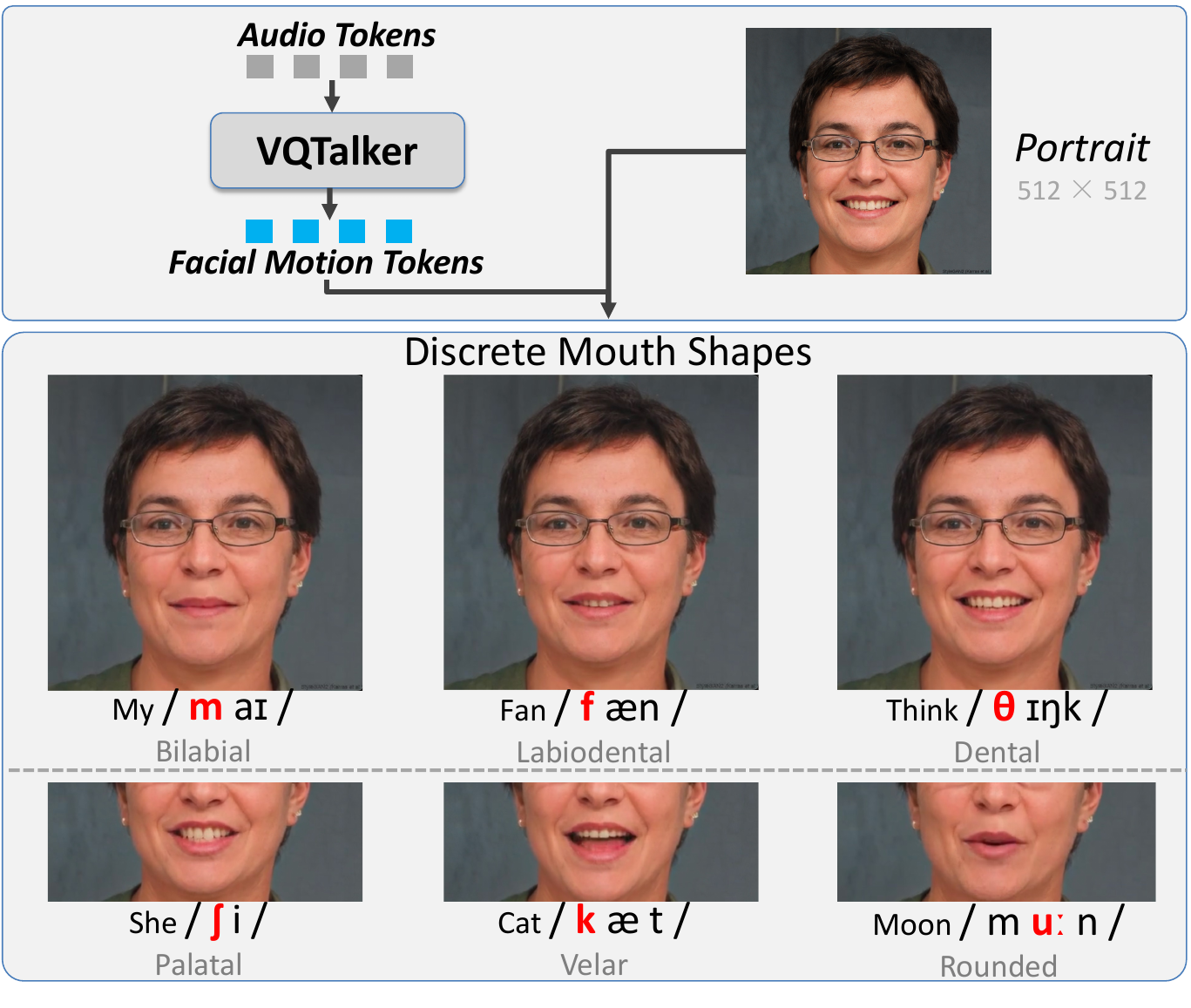}
    \caption{VQTalker converts audio tokens to facial motion tokens, animating a source portrait with diverse discrete mouth shapes.  Red text indicates phonetic sounds, with corresponding viseme categories in grey. Synthetic results can be viewed at https://x-lance.github.io/VQTalker.}
    \label{fig:viseme_vertical}
\end{figure}

In this paper, we propose \textbf{VQTalker}. Our approach leverages the fundamental linguistic concept of phonemes and visemes to create a novel system for talking head generation. By mapping discretized speech units to discretized facial motion tokens, we develop a method that captures the essential elements of speech-driven facial animation across diverse languages. Each speech unit or combination in our framework is conceptually linked to a phoneme, and each facial motion token is also conceptually tied to a viseme. This discrete representation allows us to model the universal patterns of speech-to-facial movement correlations efficiently. Figure~\ref{fig:viseme_vertical} provides a depiction of the core process by which VQTalker converts audio tokens into facial motion tokens, bringing a portrait to life with discrete mouth shapes.


Specifically, we design a facial motion vector tokenizer to model comprehensive facial movements, encompassing head pose, gaze direction, eye blinks, and most critically, the nuanced dynamics of lip shapes. These motions, particularly the intricate variations in lip articulation, form a finite set of possible states. However, directly modeling facial motion at high resolution presents challenges, primarily due to the limited capacity of a single codebook to capture complex facial dynamics. To address this, our implementation employs a multi-strategy quantization method based on Group Residual Finite Scalar Quantization~(GRFSQ). This combined approach leverages the strengths of different strategies: group quantization reduces codebook size, residual quantization allows for modeling actions at different granularities in a coarse-to-fine manner, and Finite Scalar Quantization~(FSQ) is codebook-free and improves codebook utilization. Finally, this enables us to model $512 \times 512$ resolution motion at a low bitrate of approximately 11 kbps, which is only about 70\% of the current lowest bitrate of 16 kbps for continuous representation methods~\cite{lia,anitalker}.

Building on this structure, we introduce an interleaving pattern for codebook generation that incorporates both autoregressive and non-autoregressive elements. Within each motion granularity, we employ a non-autoregressive approach to maintain bidirectional attention, fully leveraging temporal correlations. Between different granularities, we utilize an autoregressive strategy, resulting in a coarse-to-fine motion generation process. This approach strikes a trade-off between generation speed and complexity, ensuring both efficiency and quality in the produced animations.

The main contributions of our work are:
\begin{enumerate}
\item A facial motion tokenizer that leverages the finite nature of speech articulation to create a discretized representation, enhancing the capture of subtle lip movements and improving cross-lingual generalization.
\item A Group Residual Finite Scalar Quantization (GRFSQ) method that combines group, residual, and finite scalar quantization strategies to achieve efficient high-resolution facial motion modeling at low bitrates.
\item A coarse-to-fine motion generation process combining non-autoregressive temporal and autoregressive granular approaches, enhancing both consistency and efficiency.
\end{enumerate}

\section{Related Works}

\textbf{Latent Facial Representation.} Recent studies~\cite{gaia, vasa1, drobyshev2024emoportraits} have explored more refined representations through self-supervised learning. While this approach models facial features more precisely than predefined structured features like landmarks, it often implicitly encodes target identity information, leading to identity leakage issues. Although various strategies~\cite{megaportraits, anitalker} have been proposed to address this issue, they introduce additional model complexity and computational overhead. In this paper, we propose exploring \textbf{discrete coding} in Latent Facial Representation. Discrete coding inherently creates an information bottleneck by quantizing the continuous feature space into a finite discrete codebook. This approach enables the model to learn more compact representations and mitigates information leakage without additional modules or losses.

\textbf{Facial Tokenization.} Existing approaches~\cite{xing2023codetalker, tan2024flowvqtalker, yu2024image} typically rely on VQ-VAE~\cite{vqvae}, \textbf{patch-based} tokenization, which limits to local regions or specific patterns~\cite{codetalker, saas} and fails to capture global semantics, constraining their application to lower resolutions. Scaling these methods to higher resolutions often requires complex networks with billions of parameters~\cite{emo, echomimic}. We argue that the drawback lies in coupling texture and motion information. Due to the texture part containing detailed information, the network's reconstruction of these details leads to an increase in network parameters, making it difficult to scale to higher resolutions or requiring larger networks. In contrast, our method decouples these elements by independently quantizing motion and using a separate rendering network to recover the detailed parts. This approach allows us to transcend local patch semantics to \textbf{global motion semantics}, enabling holistic modeling of facial dynamics without the need for excessively large networks.

\textbf{Vector Quantization (VQ)} maps continuous input vectors to a finite set of discrete vectors for compressing data or improving the network efficiency. \textbf{Group VQ~(GVQ)}~\cite{vq_wav2vec, gvq} utilizes multiple groups to model different patterns, alleviating mode collapse in single codebooks. \textbf{Residual VQ (RVQ)}~\cite{soundstream, musicgen} quantizes vectors iteratively, encoding the residual error, which is particularly suitable for features with different granularity representations. For instance, in audio processing, shallow semantic features and deep acoustic features can be modeled at different granularities in various residual layers. In terms of codebook learning, compared with original codebook optimization, \textbf{Finite Scalar Quantization (FSQ)}~\cite{fsq} optimizes the table lookup process and incorporates the Straight-Through Estimator (STE) technique into the network. It projects the continuous representation of neural networks onto a few dimensions, then quantizes each dimension to a finite set of values. This approach achieves good performance and better codebook utilization without complex auxiliary losses. Our method will combine the strengths of those methods.

\begin{figure*}[!ht]
    \centering
    \includegraphics[width=1.0\linewidth]{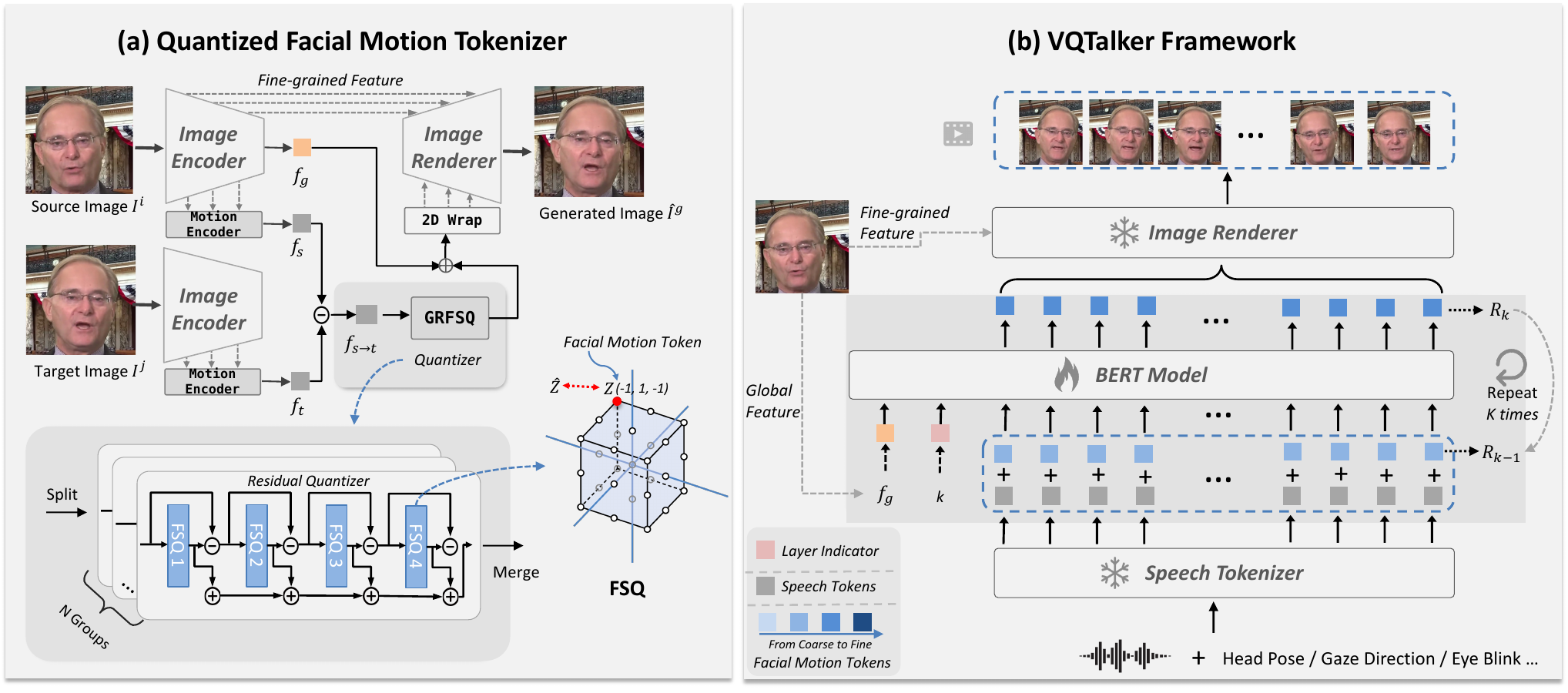}
    \caption{The VQTalker framework consists of two main components: (1) A quantized facial motion codec that learns a universal motion representation through self-supervised learning and Group Residual FSQ (GRFSQ). (2) A coarse-to-fine motion generation process using a BERT model, which takes speech tokens and iteratively generates facial motion tokens across multiple residual layers. The Image Renderer then synthesizes the final talking head video from the generated discrete codewords.}
    \label{fig:arch_main}
\end{figure*}

\section{VQTalker Framework}

Our proposed VQTalker framework addresses the challenges of multilingual talking head generation by transforming speech tokens into visual motion tokens. As illustrated in Figure~\ref{fig:arch_main}, the framework consists of two primary components: (a) a quantized facial motion tokenizer, and (b) a coarse-to-fine motion generation process.

\subsection{Quantized Facial Motion Tokenizer}

The core of our approach is a quantized facial motion tokenizer designed specifically for talking head tasks. This tokenizer, depicted in Figure~\ref{fig:arch_main}~(a), leverages self-supervised learning and a multi-strategy quantization method to create a discretized representation of facial motion, encompassing lip movements, blinking, head pose, and gaze direction.

The tokenizer follows an Encoder-Quantizer-Decoder architecture. The Image Encoder extracts three types of features from both source and target images: fine-grained facial features for high-frequency detail, global facial features $f_g$ serving as a facial prompt, and motion features $f_s$ and $f_t$ from source and target images respectively. The Quantizer converts the motion difference $f_{s\rightarrow t} = f_t - f_s$ into discrete tokens using Group Residual Finite Scalar Quantization (GRFSQ), acting as a strong information bottleneck.

The Image Renderer (Decoder) reconstructs facial features from the quantized representation, combining them with the global face feature $f_g$ to guide the final rendering. To recover fine details, each decoder layer is connected to its corresponding encoder layer through skip connections, generating 2D warps at each stage.

\begin{algorithm}[t]
\DontPrintSemicolon
\caption{Group-Residual FSQ}\label{algo:grvq_fsq}
\SetKwInput{Input}{Input}
\SetKwInput{Output}{Output}
\SetKwFunction{FSQ}{FSQ}

\Input{$\boldsymbol{x}$, FSQ levels $[l_1, l_2, ..., l_d]$, number of groups $G$, number of residual quantizers $R$}
\Output{quantized $\boldsymbol{\hat{x}}$, codebook indices $\boldsymbol{I}$}

\tcc{Split into G groups}
split $\boldsymbol{x}$ into $G$ groups: $\boldsymbol{x}_1, \boldsymbol{x}_2, \ldots, \boldsymbol{x}_G$\;

$\boldsymbol{\hat{x}}_g \gets 0$ for $g = 1$ to $G$\;
$\boldsymbol{I} \gets$ empty tensor of shape $(G, R, *)$\; 

\For{$g=1$ to $G$}{
    ${\rm residual} \gets \boldsymbol{x}_g$\;
    
    \tcc{Residual quantizers}
    \For{$r=1$ to $R$}{
        $z_{quantized}, {\rm indices} \gets$ \FSQ{${\rm residual}$}\;
        
        $\boldsymbol{\hat{x}}_g \pluseq z_{quantized}$\;
        ${\rm residual} \minuseq z_{quantized}$\;
        $\boldsymbol{I}[g, r] \gets {\rm indices}$\;
    }
}

$\boldsymbol{\hat{x}} = concat(\boldsymbol{\hat{x}}_1, \boldsymbol{\hat{x}}_2, \ldots, \boldsymbol{\hat{x}}_G)$\;

\textbf{return} $\boldsymbol{\hat{x}}$, $\boldsymbol{I}$
\end{algorithm}

\subsubsection{Group Residual Finite Scalar Quantization (GRFSQ)}

Our GRFSQ quantizer combines Group Vector Quantization (Group VQ), Residual Vector Quantization (Residual VQ), and Finite Scalar Quantization (FSQ). The GRFSQ process is visually illustrated in Figure~\ref{fig:arch_main}~(a), with the detailed algorithm presented in Algorithm~\ref{algo:grvq_fsq}.

Specifically, the input \(\boldsymbol{x}\) is first split into \(G\) groups. Each group undergoes \(R\) iterations of residual quantization, with FSQ applied to the residual at each step. Similar to the single-codebook VQ~(Vector Quantization) method, FSQ projects the continuous representation onto a fixed codeword. However, unlike traditional VQ, FSQ does not involve a codebook lookup process. Instead, each dimension is quantized to a set of fixed levels \([l_1, l_2, \ldots, l_d]\), and this process can be optimized using the Straight-Through Estimator (STE). This approach creates an implicit codebook from the product of these level sets, enabling efficient quantization without an explicit codebook.

Finally, the whole process generates \(G \times R\) sets of codebook indices \(\boldsymbol{I}\). The quantized outputs from all groups are merged to form the final quantized vector \(\boldsymbol{\hat{x}}\). These indices \(\boldsymbol{I}\) will serve as target values in the subsequent Coarse-to-Fine Motion Generation process. 

This combined strategy of the codebook offers several advantages: it enhances codebook utilization, mitigates codebook collapse without requiring complex auxiliary losses, and provides a flexible framework for capturing multi-scale facial motion features. 

\begin{figure}[h]
\centering
\includegraphics[width=0.5\textwidth]{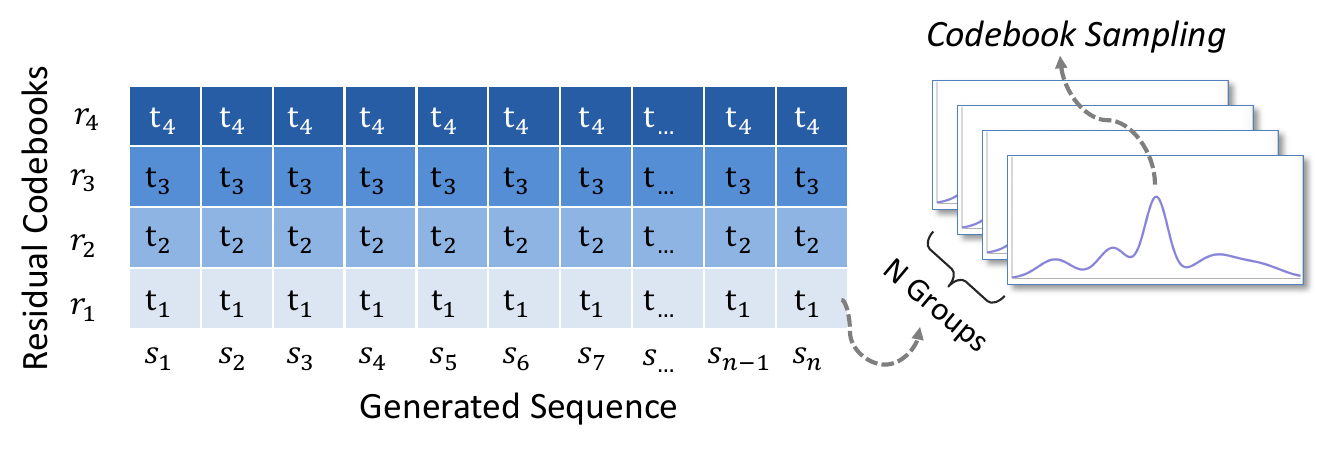}
\caption{Codebook interleaving patterns. Hybrid facial motion generation matrix. Rows ($r_1$ to $r_4$) represent residual codebooks of increasing detail. $t_1$ to $t_4$ indicate the sequential processing steps for each residual layer, with each $t_i$ generating the corresponding $r_i$ in one pass.}
\label{codebook_interleaving}
\end{figure}

\subsection{Coarse-to-Fine Motion Generation}

Our coarse-to-fine motion generation process, illustrated in Figure~\ref{fig:arch_main}~(b), transforms audio input into facial animation output. This approach employs a BERT~\cite{bert} model to convert discrete audio features into facial motion tokens, progressively refining from coarse to fine motions.

The process begins with inputs from multiple sources: (1) discrete speech tokens~($\mathbf{A}$) from a pretrained Speech Tokenizer, (2) additional control information including frame-level head pose~($\mathbf{h}$), gaze direction~($\mathbf{g}$), and eye blink~($\mathbf{b}$), (3) the global feature $f_g$, and (4) a layer indicator~($k$). These inputs are combined to form the composite input vector for each residual layer $r$:

\[\mathbf{C}_r = \{f_g, k, [\mathbf{A}, \mathbf{h}, \mathbf{g}, \mathbf{b}, \mathbf{R}_{r-1}]\}\]

where $\mathbf{R}_{r-1}$ represents the output from the previous iteration (replaced with a zero matrix when $r = 0$), and the brackets represent frame-by-frame concatenation.

The BERT model processes this input to generate facial motion tokens across $K$ residual layers, represented by the `Repeat K times' loop in Figure~\ref{fig:arch_main}~(b). Each layer refines the motion representation, progressing from coarse to fine details. The Image Renderer then synthesizes the final talking head video from these representations.

Figure~\ref{codebook_interleaving} illustrates our hybrid facial motion generation matrix, which combines non-autoregressive temporal generation with autoregressive granular refinement. Rows ($r_1$ to $r_4$) represent residual codebooks of increasing detail, while $t_1$ to $t_4$ indicate the sequential processing steps for each residual layer. Temporally (horizontally), each sequence $s_1, s_2, \ldots, s_n$ is generated simultaneously. Granularly (vertically), generation progresses through four residual layers, with each $t_i$ generating the corresponding $r_i$ in one pass.

At each sequence position and residual layer, we apply group quantization, dividing the representation into $N$ groups. During training, each group is treated as a separate classification task. At inference, we perform argmax sampling for each group. The learning objective is to reconstruct the categories of the $(r+1)$-th codebook, conditioned on the input $\mathbf{C}_r$ from the $r$-th layer, by minimizing the negative log-likelihood:

\[L(\mathcal{G}|\mathbf{C}_r) = - \sum_{g=1}^G \sum_{t=1}^T \log p_{\mathcal{G}}(\boldsymbol{I}_{g,t,r+1} | \mathbf{C}_r)\]

where $\boldsymbol{I}_{g,t,r+1}$ represents the target codeword index in the $(r+1)$-th layer codebook for the $g$-th group at time step $t$. These indices are derived from the GRFSQ process described earlier, serving as the target values for our coarse-to-fine generation. The generation across different residual layers shares the same set of parameters, distinguished by the layer indicator $k$.

This coarse-to-fine approach allows our model to capture both global facial movements and fine-grained details, resulting in more natural and expressive talking head across diverse linguistic contexts. By using the quantized indices $\boldsymbol{I}$ as targets, we maintain a direct link between the GRFSQ quantization and the motion generation process, ensuring that the generated motions accurately reflect the discretized facial dynamics captured in the tokenization stage.

Empirically, we observe that different residual layers model distinct aspects of facial motion. The first layer typically captures large-scale movements such as head pose and overall facial orientation. Subsequent layers progressively refine the motion, with middle layers focusing on more localized movements like eyebrow raises, jaw motions, and lip movements. The final few layers capture the most subtle details, including gradually reducing visual jitter. This hierarchical representation allows our model to efficiently capture the full spectrum of facial dynamics, from global head movements to micro-expressions, while also alleviating the complexity of the generation process. As a result, it contributes to the naturalness and expressiveness of the generated animations across various linguistic contexts. For a visual demonstration of this coarse-to-fine progression, we encourage readers to view the accompanying video in the supplementary materials.

\section{Experiment}

\subsection{Experiment Setup}

\textbf{Training Dataset}
We utilized three publicly available datasets: VoxCeleb~\cite{voxceleb}, HDTF~\cite{hdtf}, and VFHQ~\cite{vfhq}. To ensure consistency, we re-downloaded the original videos and processed them using a unified approach. This approach included filtering out faces smaller than 512 pixels. Our effort resulted in a dataset comprising around 16k video clips with a total duration of 210 hours. After statistical analysis, \textbf{Indo-European} languages (such as English, German, and French) account for around 97\% of the final dataset. 

\textbf{Evaluation Dataset} To evaluate the effectiveness of non-talking head algorithms on non-Indo-European languages, we compiled the \textbf{M}ultilingual \textbf{N}on-Indo-European \textbf{T}alking Head \textbf{E}valuation Corpus (\textbf{MNTE}). This dataset encompasses Arabic, Swahili, Mandarin, Korean, Japanese, and Turkish, aiming to cover a wide range of commonly used non-Indo-European language families. For each language, we collected 5 talking head video clips, each no shorter than 3 seconds, resulting in a total of 30 videos. To evaluate performance in Indo-European languages and video reconstruction tasks, we use HDTF~\cite{hdtf} as our test set, which follows Dinet~\cite{dinet}.

\textbf{Evaluation Metric} We employ a suite of measures to assess the quality and similarity of generated images and videos across different scenarios. These metrics include: Image Similarity Metrics - Structural Similarity Index (SSIM)\cite{ssim} and Learned Perceptual Image Patch Similarity (LPIPS)\cite{lpips}, which quantify structural and perceptual similarity between generated and ground truth images; Distribution-based Metrics - Fréchet Inception Distance (FID) and Fréchet Video Distance (FVD); Facial Fidelity Metrics - Cosine Similarity (CSIM), which evaluates facial similarity, and Landmark Distance (LMD), which measures key facial positioning; Video Quality Metrics - Motion Stability Index (MSI)\cite{stableface}, which assesses motion stability; and Image Quality Metrics - Cumulative Probability of Blur Detection (CPBD)\cite{cpbd}. To ensure a fair comparison, the resolution is standardized to $256 \times 256$.

\textbf{Model Configuration}
For training the face motion tokenizer, our training paradigm is primarily based on LIA~\cite{lia} and extends it to support a resolution of $512 \times 512$ pixels. Then, based on it, we add a vector quantization module. In the second stage, we employed a 12-layer BERT~\cite{bert} network to iteratively generate a four-layer residual codebook for the face tokenizer. The maximum length is 4096. A pre-trained speech tokenizer from CosyVoice~\cite{du2024cosyvoice} serves as the audio feature encoder, incorporating a downsampling layer to adjust the audio sampling rate from 50 Hz to 25 Hz to synchronize with the video frame rate. 

\begin{table}[htpb]

  \caption[Dataset statistics]{Video Reconstruction Result on HDTF. {\em Reso.} indicates image resolution (width and height), {\em Bitrate} is measured in kbps. \textbf{Bold} and \underline{underlined} values represent the best and second-best results respectively.}
  \label{tab:video_reconstruction_result}
  \centering
  \resizebox{1.0\linewidth}{!}{
  \begin{tabular}{@{}l cc|cccc@{}} \toprule
  Method & Reso. & Bitrate & SSIM $\uparrow$  & LPIPS$\downarrow$ & CSIM$\uparrow$ & CPBD$\uparrow$ \\  \cmidrule{1-7} 

H.264 Codec  & 512 & $\approx$ 347 & - &	 - & - & -  \\ 
\cmidrule{1-7}
FOMM & 256 & 48 & 0.775 &	 0.178 & 0.830 & 0.339  \\ 
DPE &256& 16 & 0.861  &	0.151 &	 0.912  &   0.354 \\ 
MTIA  &256& 48 & 0.870 &	0.122 & \textbf{0.929} & 0.316 \\ 
Vid2Vid &256& 36  & 0.870 &	 0.115 &	0.924 & 0.386\\ 
LIA &256&\underline{16} & 0.831 &	0.137 &	0.916 & 0.297 \\ 
FADM &256& 36 & 0.849    &	 0.147  &	0.916 &  0.332 \\ 
AniTalker  &256& \underline{16} &	\textbf{0.905}	& \textbf{0.079} &	\underline{0.927} &  0.367 \\ 
LivePortrait  & 512&  $\approx$  50  &	  0.800 	&   0.134  & 0.883  & 0.374\\

EMOPortrait  & 512&  $\approx$  102  &	   0.736 	&   0.206   & 0.657  &  \textbf{0.434} \\%

\cmidrule{1-7} 

\textbf{VQTalker (Ours)}  &\textbf{512}& $\approx$ \textbf{11} &	\underline{0.874} 	& \underline{0.083} &	 0.919 &  \underline{0.421} \\ 

    \bottomrule
  \end{tabular}
}
\end{table}

\subsection{Facial Motion Tokenizer}

We conducted a comprehensive evaluation of our approach against several state-of-the-art face reenactment methods~\cite{fomm, dpe, mtia, facevid2vid, lia, fadm, anitalker, echomimic, liveportrait} in a video reconstruction scenario. This scenario tests the algorithms' ability to accurately reconstruct the original video by using a video of the same identity to drive a portrait, with the first frame serving as the source image. All compared methods utilize variations of self-supervised learning. Table~\ref{tab:video_reconstruction_result} presents the results of this evaluation.

Our method, VQTalker, stands out by achieving competitive results while supporting a higher resolution ($512 \times 512$) and maintaining a significantly lower bitrate. Specifically, our approach employs 12 group layers, 4 residual layers, and 625 codebook entries per group, with a sampling rate of 25 fps, resulting in a total bitrate of approximately 11 kbps (calculated as $4 \times 12 \times \log_2(625) \times 25 $). For other methods using continuous variables, we converted them to bits by multiplying by 32-bit float, with all bitrates calculated at 25 fps. For example, LIA, which uses 20 dimensions for intermediate latent, has a bitrate of $20 \times 32 \times 25 = 16000$ (16kbps).


Despite the low bit rate, our method demonstrates strong performance in key metrics. VQTalker achieves the second-highest scores in image structural metrics such as SSIM and LPIPS. Our CSIM is very close to the state-of-the-art, demonstrating robust capabilities in preserving facial identity during reconstruction. In particular, our method achieves the highest CPBD, indicating a better image sharpness compared to other approaches. Although we did not achieve the best results in all metrics, we attribute this to information loss from quantization. For example, subtle position shifts that are imperceptible to the human eye may be captured by these metrics. Compared to traditional video codecs like H.264~\cite{wiegand2003overview}, which requires an average bitrate of approximately 347 kbps at 512 resolution, our approach achieves comparable performance while reducing the bitrate to approximately 11 kbps. This suggests that our discrete representation captures facial dynamics more efficiently than the continuous representation. For 512-resolution algorithms, LivePortrait~\cite{liveportrait} requires a bitrate of approximately 50 kbps, but achieves lower performance in most metrics compared to our method. These comparisons highlight our ability to maintain high-quality outputs at higher resolutions while reducing the bandwidth.

\begin{table*}[htpb]
  \caption[Dataset statistics]{Quantitative comparisons with speech-driven baselines on the HDTF (Indo-European) and MNTE (Non-Indo-European) datasets. \textbf{Bold} and \underline{underlined} values represent the best and second-best results respectively.}
  \label{tab:audio_driven_result}
  \centering
    \resizebox{1.0\linewidth}{!}{
  \begin{tabular}{@{} l cccccccccccc@{}} \toprule
  \multirow{2}{*}{Method} 
     &\multicolumn{6}{c}{HDTF~(Indo-European)} & \multicolumn{6}{c}{ MNTE~(Non-Indo-European)}  \\  \cmidrule(lr){2-7}  \cmidrule(lr){8-13} 
                 & SSIM$\uparrow$  & CSIM$\uparrow$ & LMD$\downarrow$  &  CPBD$\uparrow$ &FID$\downarrow$ & FVD$\downarrow$ &SSIM$\uparrow$  & CSIM$\uparrow$ & LMD$\downarrow$  &  CPBD$\uparrow$ & FID$\downarrow$  &FVD$\downarrow$ \\  \cmidrule{1-13} 
SadTalker~\cite{sadtalker}  &	0.510 & 0.726 & 0.320 & 0.391 & 37.699 & 526.194   &  0.456 & 0.423 & 0.686& 0.216 &  81.128 & 564.047  \\ 
EAT~\cite{eat}  & 	0.433 & 0.678 & 0.388 &0.334 & 177.317 & 667.499 & 0.394 & 0.420 & 0.663 & 0.224 & 158.465 & 540.128 \\ 
PD-FGC~\cite{pdfgc}   &  	 0.325 & 0.298 & 0.671 & 0.264 & 214.200 & 846.211  &   0.357 & 0.302& 0.726 & 0.212& 154.394 & 624.143  \\ 
AniTalker~\cite{anitalker}   & 0.663	& 0.709   & 0.426  & \underline{0.394} & \underline{34.705} & 444.162 & 0.520 & 0.490 & 0.671 & \underline{0.246}& 65.780 & 430.209 \\
EDTalker~\cite{edtalk}   & \underline{0.823}	& \underline{0.881} & \underline{0.084} & 0.301&   36.437  & 410.709   & \textbf{0.745} & \underline{0.802} & \textbf{0.107} & 0.219  & \underline{55.070} & \underline{321.825}  \\
EchoMimic~\cite{echomimic}  &  0.690	&    \underline{0.881}  & 0.094 & 0.385& 36.354 & \underline{272.425}  &   0.600 & 0.737 & 0.213 & 0.238    & 62.185 & 391.568 \\
\cmidrule{1-13} 

\textbf{VQTalker (Ours)}  &	\textbf{0.835 } & \textbf{ 0.902 }  & \textbf{0.067}   &  \textbf{0.408 } & \textbf{28.783}  & \textbf{205.471} & \underline{0.730 }& \textbf{ 0.809} & \underline{0.146} & \textbf{0.263} & \textbf{43.411} & \textbf{233.586}  \\

    \bottomrule
  \end{tabular}
}
\end{table*}

\begin{figure*}[!pt]
    \centering
    \includegraphics[width=0.95\linewidth]{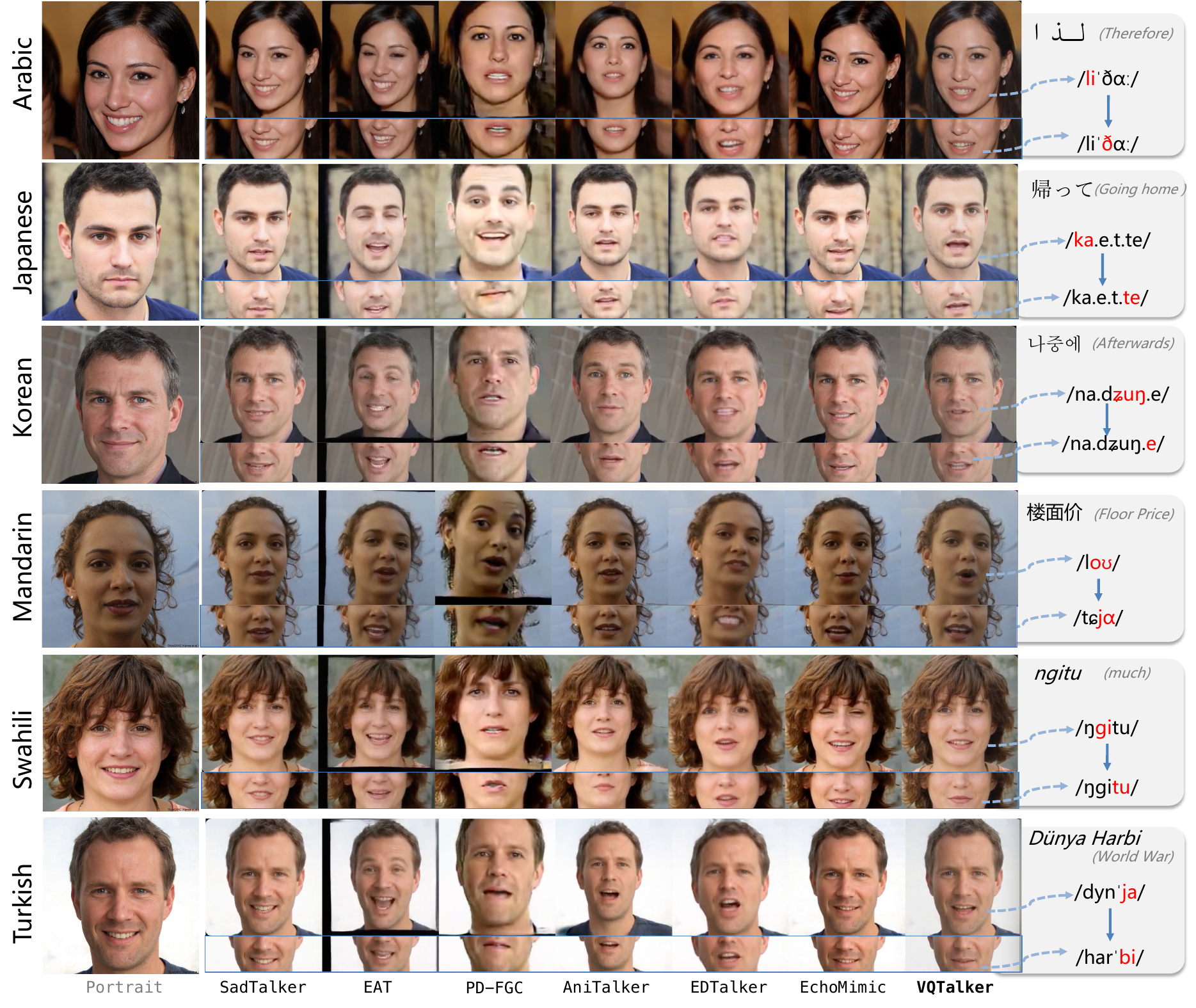}
\caption{Qualitative comparison of speech-driven talking head generation methods across non-Indo-European languages from Arabic to Turkish. Each row shows a source portrait (leftmost) animated by different algorithms to pronounce specific syllables (indicated in red). The upper and lower images in each example demonstrate the transition in lip movements. This visualization highlights our method's ability to model comprehensive pronunciation features and support multiple languages. To better observe the motion changes and co-articulation capabilities, it is recommended to watch the supplementary video.}
    \label{fig:audio_driven_demo}
\end{figure*}

\subsection{Coarse-to-fine Experiment}

We evaluate our method against several state-of-the-art speech-driven approaches: SadTalker~\cite{sadtalker}, EAT~\cite{eat}, PD-FGC~\cite{pdfgc}, AniTalker~\cite{anitalker}, EDTalker~\cite{edtalk}, and EchoMimic~\cite{echomimic}. EchoMimic operates at $512 \times 512$ resolution, while the other methods work at $256 \times 256$. Our experimental setup involves using the audio from a video to animate its first frame, aiming to recreate the original video as accurately as possible. This evaluation is conducted across both Indo-European and non-Indo-European languages to assess cross-lingual performance.

The quantitative results, presented in Table~\ref{tab:audio_driven_result}, demonstrate the superior performance of VQTalker. Our method achieves top scores in identity preservation (CSIM), image sharpness (CPBD) and distribution distance (FID and FVD). Additionally, VQTalker maintains competitive results in structural similarity (SSIM) and landmark accuracy (LMD). Notably, our approach exhibits a smaller performance gap between Indo-European and non-Indo-European languages compared to other methods, underscoring its robust cross-lingual generalization. The qualitative results, illustrated in Figure~\ref{fig:audio_driven_demo}, provide visual evidence of VQTalker's effectiveness in capturing pronunciation actions across diverse non-Indo-European languages. Our method demonstrates notable accuracy in lip movements while consistently preserving identity. This is particularly evident in the natural mouth shapes and fluid transitions observed for various phonemes, such as `/ka.e.t.te/' in Japanese. These examples highlight VQTalker's proficiency in modeling a wide range of phonetic articulations across different linguistic systems. The visual comparison further emphasizes our model's ability to adapt to the unique characteristics of each language.

\subsection{Ablation Study}

\begin{table}[htpb]

  \caption[Dataset statistics]{Vector Quantization Ablation on HDTF Cross-Identity Driven: \#~$\text{G}$ (groups), \#~$\text{R}$ (residuals), FSQ (Finite Scalar Quantization), \#~$\text{Codes}$ (codebook size / FSQ levels), {\em Util.}: Codebook Utilization Rate (code used at least once)}
  \label{tab:QuantitativeVD}
  \centering
  \resizebox{1.0\linewidth}{!}{
  \begin{tabular}{@{}cccccc|ccc@{}} \toprule
  Method & \# G & \# R & FSQ & \# Codes  & Bitrate  & CSIM $\uparrow$  & MSI $\uparrow$  & Util.(\%) $\uparrow$ \\  \midrule
VQ & 1 & - & - & 8196 & $\textless 1$ &  0.588   & 0.588 & 0.35  \\ 
GVQ & 32 & - & -  & 1024 & 8 & 0.548  &  0.052 & 88.89 \\ 
RVQ & 1 & 32 &  - & 1024 & 8 & 0.561  & 0.473 &  50.00\\ 
GRVQ & 12 & 4 &  - & 1024 &  12  &  0.572 & 0.359 & 26.78\\ 

\midrule
GRFSQ & 8 & 4 & \checkmark &  $5\times5\times5\times5$ & $\approx 7$  & 0.576& 1.016 & 73.99\\ 
GRFSQ & 12 & 2 & \checkmark & $5\times5\times5\times5$ & $\approx 6$ & 0.576  & 0.616 & 56.89\\ 

GRFSQ & 12 & 4 & \checkmark & $4\times4\times4\times4$ & 9.6  &  0.570 &  0.656 & 59.68\\ 
GRFSQ & 12 & 4 & \checkmark &  $6\times6\times6\times6$ &  $\approx 12$  &  0.571 & 0.932 &  65.64\\ 

\midrule 

\textbf{GRFSQ}  & 12 & 4 &  \checkmark &  $5\times5\times5\times5$  &	 $\approx 11$ & 0.571  & 0.992 &  78.74\\ 

    \bottomrule
  \end{tabular}
}
\end{table}

\subsubsection{Codebook Design}
Table~\ref{tab:QuantitativeVD} compares different codebook designs in cross-identity-driven scenarios. We focus on cross-identity scenarios here as they reveal more significant differences between methods, whereas self-driven scenarios show less pronounced variations. Classic designs (VQ, GVQ, RVQ, GRVQ) generally show lower MSI scores, indicating increased jitter on the generated result, suggesting limited model capacity in modeling complex facial dynamics. Our GRFSQ variations demonstrate that balancing group number, residual layers, and FSQ levels is crucial. The optimal configuration (12 groups, 4 residuals, $5 \times 5 \times 5 \times 5$ FSQ levels) achieves a good trade-off between CSIM, MSI, and codebook utilization. Notably, GRFSQ outperforms GRVQ in MSI (0.992 vs 0.359) and utilization rate (78.74\% vs 26.78\%), suggesting FSQ's structure allows for more efficient use of the representation space, potentially leading to improved motion modeling and stability.

\begin{table}[htpb]

  \caption[Dataset statistics]{Ablation on Discrete (D) and Continuous (C) Representation. X-Y format: X for input (Audio Feature), Y for output (Motion Feature).}
  \label{tab:ContinuousFeat}
  \centering
    \resizebox{1.0\linewidth}{!}{
  \begin{tabular}{@{}lcc|cccc@{}} 
     \toprule
  Method   & Audio Feature & Motion Feature  & SSIM$\uparrow$ & CSIM$\uparrow$ & LPIPS $\downarrow$  & CPBD$\uparrow$  \\  
  \midrule
C-C & Whisper & Vector   & 0.795 & 0.889 &0.101 & \textbf{0.410} \\
D-C & CosyVoice  & Vector  &  0.769 & 0.882 & 0.115 &\textbf{0.410}  \\ 
C-D & Whisper & GRFSQ   &\underline{0.821}& \underline{0.890} & \underline{0.085} & \underline{0.408 }   \\
D-D & VQ-Wav2vec  & GRFSQ   &	 0.804 & 0.887 & 0.089 & \underline{0.408}    \\
\midrule
\textbf{D-D} & CosyVoice  & GRFSQ   &  \textbf{0.835} & \textbf{0.902} & \textbf{0.067} & \underline{0.408}   \\ 
    \bottomrule
  \end{tabular}
}
\end{table}

\subsubsection{Discrete vs. Continuous Representation}
Table~\ref{tab:ContinuousFeat} compares discrete and continuous representations for audio and motion features. For continuous audio representation, we employed the multilingual Whisper Large v2~\cite{radford2023robust} model, ensuring a fair comparison across different languages. Discrete representations consistently outperform continuous ones, especially when used for both input and output (D-D). Our proposed method (CosyVoice for audio, GRFSQ for motion) achieves the best overall performance, with notable improvements in SSIM (0.835) and LPIPS (0.067). This suggests that discrete representations can more effectively capture relevant information for both audio and motion features, even when compared to advanced multilingual models like Whisper. Among speech tokenizers, CosyVoice outperforms VQ-Wav2vec~\cite{vq_wav2vec}, which may be attributed to its more compact and efficient codebook design. These results underscore the potential of leveraging discrete representations in both audio and motion domains for multilingual talking head generation tasks.

\section{Conclusion}

We introduced VQTalker, a framework designed for multilingual talking head generation, utilizing discretized facial motion representations. The Group Residual Finite Scalar Quantization (GRFSQ) method within VQTalker successfully trades off low bitrate with high-quality facial animation. VQTalker particularly excels in speech-driven scenarios, offering robust performance across non-Indo-European languages. The coarse-to-fine generation approach contributes to improved temporal consistency and natural lip synchronization, even in diverse linguistic contexts. These results underscore the effectiveness of discretized representations in capturing complex facial dynamics while maintaining efficiency, advancing the development of more inclusive and versatile talking head technologies.

\clearpage

\appendix

\section{Supplementary Materials}

\section{Terminology}

To avoid confusion, this section provides descriptions of some terminology that may lead to ambiguity.

\textbf{Tokenizer and Quantizer.} In this paper, we use the term `tokenizer' to refer to a process that converts continuous facial motion data into a discrete representation. Our facial motion tokenizer encompasses both the extraction of facial features and their quantization into a finite set of discrete tokens. This process includes a quantization step, which we specifically refer to as a `quantizer' when discussing the technical details of our Group Residual Finite Scalar Quantization (GRFSQ) method.

\textbf{Tokens vs. Discrete Codeword.}  In this paper, we use the terms `tokens' and `discrete codewords' to describe the discrete representations derived from continuous data. While these terms can be used interchangeably in the context of our methodology, they have slightly different connotations: `Tokens' generally refer to the abstract units of representation that the model uses to represent audio and facial motion information. They are the conceptual building blocks that our model works with after processing the input data. `Discrete codewords', on the other hand, are the specific quantized values that result from applying the quantization process to continuous input data. They represent the actual, concrete output of our discretization step.

\textbf{Phonetic Classifications.} Phonetic classifications refer to the categorization of speech sounds based on how they are produced by the vocal organs, including the position and movement of the lips, tongue, and other articulators. While phonetic classifications in English encompass numerous categories (including bilabial, labiodental, dental, alveolar, sibilant, palatal, velar, rounded, unrounded, open, neutral, and retroflex)~\cite{davenport2020introducing}, Figure 1 presents only six discrete mouth shapes: bilabial, labiodental, dental, palatal, velar, and rounded. These shapes were selected because they are easily distinguishable in static images. For a more comprehensive understanding of additional mouth shapes and coarticulation effects, we recommend observing our demo videos.

\textbf{Patch-based VQ vs. Global Semantic VQ.} Existing image VQ methods~\cite{vqvae} predominantly rely on local, patch-based approaches. In these methods, an image is typically divided into small, non-overlapping patches, each of which is then independently quantized. This approach, while effective for general image compression, may not be optimal for facial motion representation. For human faces, we argue that considering global semantics is crucial for capturing comprehensive facial dynamics. Our method adopts a global semantic VQ approach, which quantizes the entire facial motion representation as a whole, rather than in patches. This global perspective allows for better capture of holistic facial expressions, head movements, and inter-feature relationships. Moreover, it potentially reduces model complexity by eliminating the need for numerous local codebooks. By focusing on global facial semantics, our approach can more efficiently represent complex facial motions while maintaining lower computational requirements.

\section{Dataset Details}

\subsection{Training Dataset}

Our data collection pipeline contains three distinct stages, utilizing the datasets VoxCeleb~\cite{voxceleb}, HDTF~\cite{hdtf}, and VFHQ~\cite{vfhq}. As VFHQ does not include audio tracks, this dataset is only used in the first stage of training. For all these datasets, our unified processing procedure is as follows:

\begin{enumerate}
    \item \textbf{Re-downloading Original Datasets}: To ensure uniform processing, given the different data handling methods across datasets, we downloaded the original videos. For VoxCeleb and HDTF, changes in the original sources meant we could only secure about 60-70\% of the initial datasets. The VFHQ authors provided the complete set of videos, obviating the need for re-downloading.
    
    \item \textbf{Face Detection and Face Tracking}: This step involves detecting faces in videos. In contrast to previous studies, we chose not to align the faces to allow for positional shifts within the frame, aiming to preserve natural head movements.

    \item \textbf{Applying Filtering Rules}: Our filtering process involved two main criteria. We first excluded faces with resolutions lower than $512\times 512$.
    

 \item \textbf{Resize to 512 $\times$ 512}: All our images, whether for training or testing, are originally based on the resolution of 512 $\times$ 512. Therefore, the purpose of this step is to resize all images to 512 $\times$ 512.
\end{enumerate}

\begin{figure}[t]
\centering
\includegraphics[width=1.0\linewidth]{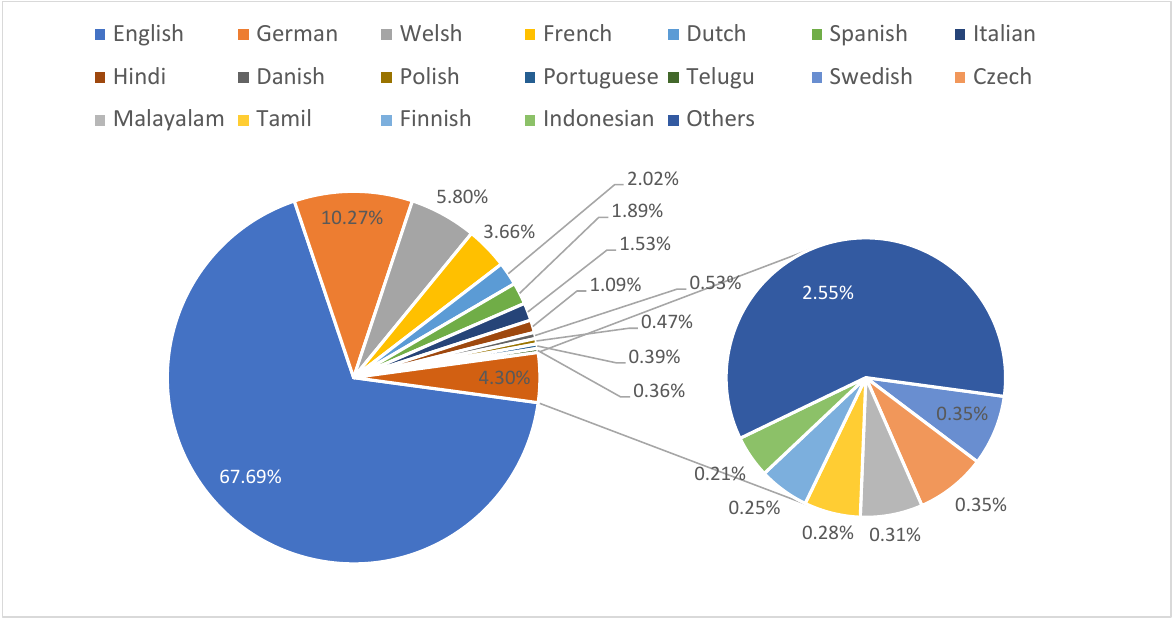}
\caption{Language distribution in our constructed training dataset. Indo-European languages constitute around 97\% of our training data.}
\label{fig:language_counting}
\end{figure}

Our efforts culminated in the creation of a substantial dataset comprising approximately 160,000 video clips, with a total duration of 210 hours. We employed the Whisper model to detect the language of each clip. Subsequent statistical analysis revealed that Indo-European languages dominate our constructed dataset, accounting for around 97\% of the content. Among these, English represents the largest proportion at approximately 68\%, followed by other Indo-European languages such as German and French. The specific distribution can be seen in Figure~\ref{fig:language_counting}.

\begin{figure}[t]
\centering
\includegraphics[width=1.0\linewidth]{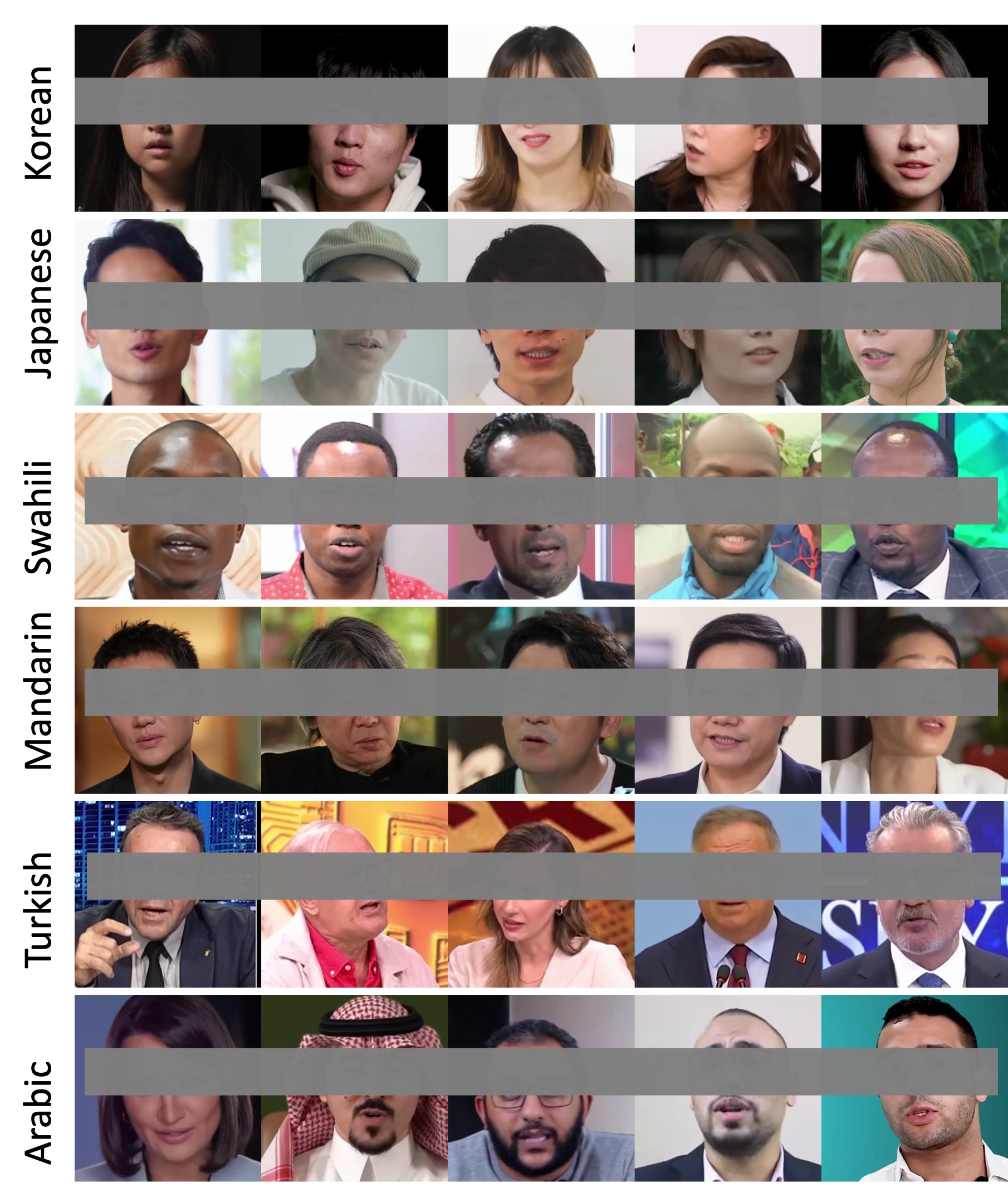}
\caption{Overview of the \textbf{M}ultilingual \textbf{N}on-Indo-European \textbf{T}alking Head \textbf{E}valuation Corpus (\textbf{MNTE}). Gray boxes are used to protect privacy.}
\label{fig:final_collage}
\end{figure}

\subsection{Evaluation Dataset}
To assess the efficacy of non-talking head algorithms on non-Indo-European languages, we curated the \textbf{M}ultilingual \textbf{N}on-Indo-European \textbf{T}alking Head \textbf{E}valuation Corpus (\textbf{MNTE}). This comprehensive dataset, sourced from the internet, encompasses a diverse range of commonly used non-Indo-European language families, including Mandarin Chinese, Korean, Japanese, Arabic, Swahili, and Turkish. As illustrated in Figure~\ref{fig:final_collage}, the corpus comprises 5 talking head video clips for each language, ensuring a robust representation of linguistic diversity. The shortest clip duration is 3.25 seconds, the average is 10.4 seconds, and the longest is 21.73 seconds. Several samples of our dataset are provided in the supplementary material.

\begin{table}[htpb]
  \caption[Dataset statistics] {Audio-driven methods and their training datasets}
  \label{tab:methods_and_datasets}
  \centering
    \resizebox{1.0\linewidth}{!}{
  \begin{tabular}{@{}l | c@{}} \toprule
Methods & Training Dataset \\  \midrule
SadTalker~\cite{sadtalker}  &	 VoxCeleb \\ 
EAT~\cite{eat}  & VoxCeleb, MEAD	\\ 
PD-FGC~\cite{pdfgc}   &  	VoxCeleb, MEAD\\ 
AniTalker~\cite{anitalker}   &  VoxCeleb, HDTF \\
EDTalker~\cite{edtalk}   &  MEAD, HDTF  \\
EchoMimic~\cite{echomimic}  &    HDTF, CelebV-HD, Interal Db \\
EMOPortraits~\cite{drobyshev2024emoportraits}  &    VoxCeleb, FEED~\cite{drobyshev2024emoportraits} \\
\textbf{VQTalker (Ours)}  &  VoxCeleb, HDTF  \\

    \bottomrule
  \end{tabular}
}
\end{table}

\begin{table}[htpb]
\caption[Dataset statistics]{Proportion of Indo-European languages in the overall datasets}
\label{tab:Proportion}
\centering
\resizebox{1.0\linewidth}{!}{
\begin{tabular}{@{}l | c@{}} \toprule
Methods & Proportion \\  \midrule
VoxCeleb~\cite{voxceleb}  &	 97.22\% \\
HDTF~\cite{hdtf}  & 100\%	\\
MEAD~\cite{mead}   &  	100\%	\\
CelebV-HQ~\cite{zhu2022celebv}   &  93.12\% \\
TalkingHead-1KH~\cite{facevid2vid}   &   93.17\%  \\
\bottomrule
\end{tabular}
}
\end{table}

\subsection{Details on Other Datasets}
To analyze the datasets used by various speech-driven baselines, we compiled statistics on audio-driven methods and their training datasets, as shown in Table~\ref{tab:methods_and_datasets}.

In addition to the VoxCeleb~\cite{voxceleb} and HDTF~\cite{hdtf} datasets, we also analyzed the proportion of non-Indo-European languages in two other frequently used facial datasets: MEAD~\cite{mead}, CelebV-HQ~\cite{zhu2022celebv}, and TalkingHead-1KH~\cite{facevid2vid}. We utilized Whisper to analyze the language distribution, and the proportion of Indo-European languages in these overall datasets can be seen in Table~\ref{tab:Proportion}\footnote{Note that these results are as of August 2024, and some videos were no longer available for download.}.

Our findings reveal that, except for some datasets that include their own internally collected data (Internal DB), the majority of datasets used in these methods predominantly feature Indo-European languages. This observation underscores the existing bias in the current corpus of speech and facial datasets.

\begin{figure}[t]
\centering
\includegraphics[width=0.5\linewidth]{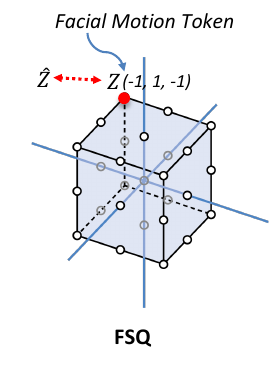}
\caption{FSQ Quantization Details}
\label{fig:fsq}
\end{figure}

\section{Experimental Details}

\subsection{Model Implementation Details}

\textbf{Stage 1: Encoder-Quantizer-Decoder} We modified the original LIA~\cite{lia} implementation by removing the Linear Motion Decomposition (LMD) component. Instead, we compute the difference between the source and target, followed by quantization. This quantized difference is then added to the direction vector. The remaining network components, including the warping module, remain unchanged. The input dimension to GRFSQ is 120, which is split into 12 groups. Each residual processing unit handles 10-dimensional data. After passing through the VQ bottleneck structure and subsequent processing, the data is merged back into a 120-dimensional output. The total number of trainable parameters in this stage is 47.7M. Once trained, this stage serves two purposes for the second stage: the encoder and quantizer are used to generate training data, while the decoder functions as the rendering network during inference.

\textbf{Stage 2 Codeword Generation:} We employ a BERT model consisting of 12 transformer layers. The concatenated input dimension is 1024, which is transformed through intermediate hidden layers of dimension 256. The final layer is a 12-layer MLP with an output dimension of 625, representing 625 classifications. This model contains 204M parameters. This model structure and size are consistently used across both discrete and continuous ablation experiments. For our GRFSQ generation, we implement an iterative generation process where different iterations share the same set of modules.
\subsection{FSQ details}

Figure~\label{fig:fsq} illustrates the Finite Scalar Quantization (FSQ) process~\footnote{https://github.com/google-research/google-research/tree/master/fsq} for facial motion tokens. In FSQ, the continuous representation $Z$ is projected onto a low-dimensional space, typically with fewer than 10 dimensions. Each dimension is then quantized to a finite set of fixed values, creating an implicit codebook. The quantization process involves bounding each dimension to a predefined range, applying a non-linear transformation (often using tanh), and then rounding to the nearest integer. This results in the quantized representation $Z_{hat}$, which in this example is mapped to the point (-1, 1, -1) in the discrete 3D space. The cube in the figure represents the possible quantization levels for each dimension, with the quantized point $Z_{hat}$ corresponding to one of the vertices of this cube. For details, please refer to FSQ~\cite{fsq}.

FSQ can be viewed as a substitute for the single VQ quantizer in our quantization process, seamlessly integrating with group and residual components. Each quantized value represents a facial motion token, which, when combined with group and residual values, forms a meaningful facial token for a complete image. The advantages of using FSQ instead of VQ include: no explicit codebook is required, no additional loss functions or modules are needed, and it allows for collaborative optimization with the network. Furthermore, the absence of an explicit codebook reduces the risk of identity leakage during the self-supervised learning process.

\subsection{Model Hyper-Parameters}

For the $L_{motion}$ loss function, we set the weighting parameters as follows: $\lambda_1 = \lambda_2 = 1000$ for eye and mouth region reconstruction losses, $\lambda_3 = 0.01$ for perceptual loss, and $\lambda_4 = 1$ for adversarial loss. These values were determined through extensive experimentation to optimize the model's performance in capturing facial motions.

\subsection{Input Feature}

In our ablation study, we evaluated two additional audio features for comparison. The first is a continuous feature extracted using the Whisper~\cite{radford2023robust} Large v2\footnote{https://github.com/openai/whisper} model as the speech encoder. This choice was motivated by two factors: Whisper provides continuous features, and it has been tested on multiple languages, allowing us to assess the impact of audio features across diverse linguistic contexts.
The second comparative discrete feature is VQ-Wav2vec~\cite{vq_wav2vec}\footnote{https://github.com/facebookresearch/fairseq}. While similar to CosyVoice~\cite{du2024cosyvoice}\footnote{https://github.com/FunAudioLLM/CosyVoice} in providing discrete audio features, VQ-Wav2vec employs self-supervised learning to obtain a robust speech representation, whereas CosyVoice is specifically trained for ASR tasks. This comparison allows us to evaluate the effectiveness of different approaches to discrete audio feature extraction. For $\mathbf{A} = {a_1, a_2, ..., a_T} \in {1, 2, ..., M}^T$, M represents the number of speech codebooks. For CosyVoice, the number is 4096, while for VQ-Wav2vec, as it is divided into 2 groups, the total number is $320 \times 320$.

For head pose ($\mathbf{h} \in \mathbb{R}^3$), we utilize 3DDFA\_V2~\cite{guo2020towards}, and for gaze direction ($\mathbf{g} \in \mathbb{R}^2$), we use a method from this link\footnote{https://github.com/hysts/pytorch\_mpiigaze\_demo}. The degree of eye blinking ($\mathbf{b} \in \mathbb{R}^2$) is obtained through a landmark predictor \footnote{https://github.com/DefTruth/torchlm}, which provides the width-to-height ratio of both left and right eyes.

\subsection{Training and Inference}
For training, we utilized four A100 (40G) GPUs, training each phase until the loss converged. Besides computing the perceptual loss, we did not incorporate any pre-trained parameters. The first phase involved initially training a model with input size $256 \times 256$, then using these parameters as initialization to train a $512 \times 512$ model. The 256 stage took 72 hours, followed by an additional 48 hours for the 512 stage. The second phase required a longer training time, approximately 120 hours. Insufficient training time can result in noticeable jitter in the rendered images. For inference, we utilized a GeForce RTX 3090 (24G) GPU and Intel(R) Xeon(R) CPU E5-2696 v3 CPU. The process begins by generating a motion sequence from the speech tokenizer, followed by frame-by-frame rendering.

\subsection{Evaluation Details}
\textbf{Scenario Setting} We evaluate methods under two distinct scenarios: speech-driven and video-reconstruction. The speech-driven scenario is designed to test the algorithms' performance under various audio-driven conditions, assessing their ability to generate appropriate facial expressions and movements in response to different speech inputs. In these tasks, when posture information is required, it is supplied from the ground truth to isolate the effect of audio on facial animation. Conversely, the video-reconstruction scenario is implemented to evaluate the effectiveness of the face tokenizer, focusing on its capacity to accurately encode and reconstruct facial features and expressions from video data.

\subsection{Video codec bitrate calculation details}

\begin{table}[htpb]

\caption[Dataset statistics]{Evaluation of Video Reconstruction on HDTF. {\em Reso.} denotes the dimensions (width x height) of the image. {\em Bitrate} is expressed in kbps. {\em Equation} describes the method used to calculate the Bitrate results. (1 k = 1000)}
  \label{tab:video_reconstruction_result}
  \centering
  \resizebox{1.0\linewidth}{!}{
  \begin{tabular}{@{}l cc|r@{}} \toprule
  Method & Reso. & Bitrate & Calculation \\  \cmidrule{1-4} 
  
H.264  & $512 \times 512$ & $\approx$ 347 & NA  \\ 
\cmidrule{1-4} 
FOMM~\cite{fomm} & $256 \times 256$ & 48 &  $ (20 + 40) \times 32 \times 25 = 48 \text{ k} $  \\ 
DPE~\cite{dpe} &$256 \times 256$& 16 & $ 20 \times 32 \times 25 = 16 \text{ k} $  \\ 
MTIA~\cite{mtia}  &$256 \times 256$ & 48 &  $ (20 + 40) \times 32 \times 25 = 48 \text{ k} $  \\ 
Vid2Vid~\cite{facevid2vid} &$256 \times 256$& 36  & $ (15\times3) \times 32 \times 25 = 36 \text{ k}$  \\ 

LIA~\cite{lia} &$256 \times 256$&\underline{16} & $ 20 \times 32 \times 25 = 16 \text{ k} $  \\ 
FADM~\cite{fadm} &$256 \times 256$& 36 & $ (15\times3) \times 32 \times 25 = 36 \text{ k}$  \\ 
AniTalker~\cite{anitalker}  & $256 \times 256$& \underline{16} &	$ 20 \times 32 \times 25 = 16 \text{ k} $  \\ 

EMOPortraits~\cite{drobyshev2024emoportraits}  & $512 \times 512$ &  $\approx$ 102 &  $ 128 \times 32 \times 25 = 102.4 \text{ k}$ \\ 

LivePortrait~\cite{liveportrait}  & $512 \times 512$ &  $\approx$ 50 &  $ (21\times3) \times 32 \times 25 = 50.4 \text{ k}$ \\

\cmidrule{1-4} 

\textbf{VQTalker (Ours)}  & $512 \times 512$& $\approx$ \textbf{11} & $4 \times 12 \times \log_2(625) \times 25 \approx 11.15 \text{ k}  $ \\ 

    \bottomrule
  \end{tabular}
}
\end{table}

To determine the bitrate of an \textbf{H.264} video, we employ FFprobe to extract the video's metadata. Within this metadata, we locate the `bit\_rate' field. To obtain a standard representation of the H.264 video's bitrate, we convert this value from bits per second to kilobits per second (kbps) by dividing by 1,000.

\textbf{FOMM}~\cite{fomm} and \textbf{MTIA}~\cite{mtia} uses 10 2D keypoints (20 dimensions) and 10 $2 \times 2$ Jacobian Matrices (40 dimensions). Each frame requires a 60-dimensional continuous vector for representation. If we assume each dimension uses a 32-bit floating-point number and there are 25 frames per second, the total number of bits needed to represent the motion of the entire frame per second is: $ (20 + 40) \times 32 \times 25 = 48,000 \text{ bits}$. This equals 48 kbps.

\textbf{DPE}~\cite{dpe}, \textbf{AniTalker}~\cite{anitalker}, and \textbf{LIA}~\cite{lia} share the same structure, so these three models have consistent dimensions. They all use 20 dimensions to represent facial actions (including expressions and poses). The calculation method is $20 \times 32 \times 25 = 16,000 \text{ bits}$. This equals 16 kbps.

\textbf{Vid2Vid}~\cite{facevid2vid}  and \textbf{FADM}~\cite{fadm} use learnable 3D keypoints as driving features. By default, it uses a total of 15 3D coordinates. The calculation method is $ (15\times3) \times 32 \times 25 = 36,000 \text{ bits}$. This equals 36 kbps.

\textbf{EMOPortraits}~\cite{drobyshev2024emoportraits} uses 128-dim vector for expression representation. The calculation method is $ 128 \times 32 \times 25 = 102,400 \text{ bits}$. This equals 102.4 kbps.

\textbf{LivePortrait}~\cite{liveportrait} uses keypoints, rotation, scale, expression, and translation (1x3) to form learnable 3D keypoints, where the learnable 3D keypoints consist of 21 3D keypoints. The calculation method is $ (21\times3) \times 32 \times 25 =  50,400 \text{ bits}$. This equals 50.4 kbps.

\textbf{VQTalker (Ours)} employs 12 group layers, 4 residual layers, and 625 codebook entries per group, resulting in a total bitrate of approximately 11 kbps (calculated as $4 \times 12 \times \log_2(625) \times 25 = 11,150$). This equals 11.15 kbps.

\begin{figure}[h]
  \centering
 \includegraphics[width=0.5\textwidth]{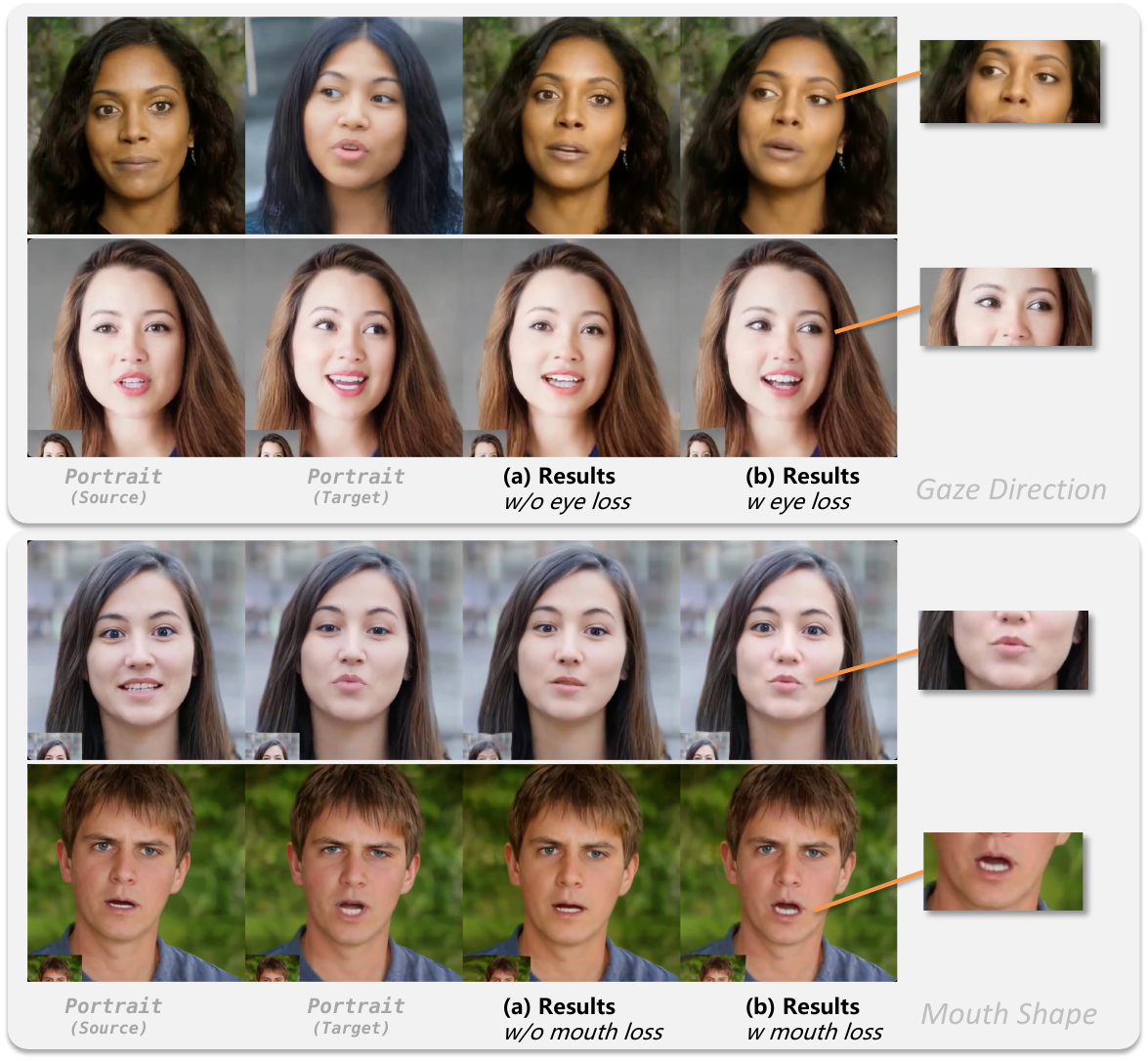}
    \caption{Ablation study on eye gaze and mouth shape modeling. Comparison of results with and without specific losses for eye and mouth regions, demonstrating improved facial feature control without additional specialized models. Image source: VASA-1\cite{vasa1}}
\label{loss_function_abalation_study}
\end{figure}

\section{Supplementary Experiments}

\subsection{Effects of Loss Weight}

In this section, we conduct ablation experiments on modeling eye and mouth movements to demonstrate the effectiveness of our region-specific weighting approach. Figure~\ref{loss_function_abalation_study} illustrates the results of these experiments.

The top two rows show gaze modeling examples, comparing results with $\lambda_1 = 0$ (column a) and $\lambda_1 = 1000$ (column b). The bottom two rows demonstrate mouth shape modeling, with $\lambda_2 = 0$ (column a) and $\lambda_2 = 1000$ (column b). 

Our results clearly show improvements in eye movement and mouth shape when the respective losses are applied. With $\lambda_1 = \lambda_2 = 1000$, the eyes more accurately follow the target gaze, and the mouth shapes more faithfully reproduce the target expressions compared to when these weights are set to 0.

These experiments demonstrate that our method can enhance the modeling of specific facial features without introducing additional pre-trained models, such as the gaze model~\cite{cortacero2019rt} used in MegaPortrait~\cite{drobyshev2022megaportraits} or the Wing Loss~\cite{feng2018wing} employed by LivePortrait~\cite{liveportrait}. By simply applying higher weights to the eye and mouth regions, we can improve the network's ability to model gaze direction and lip movements.

\begin{figure}[h]
  \centering
 \includegraphics[width=0.5\textwidth]{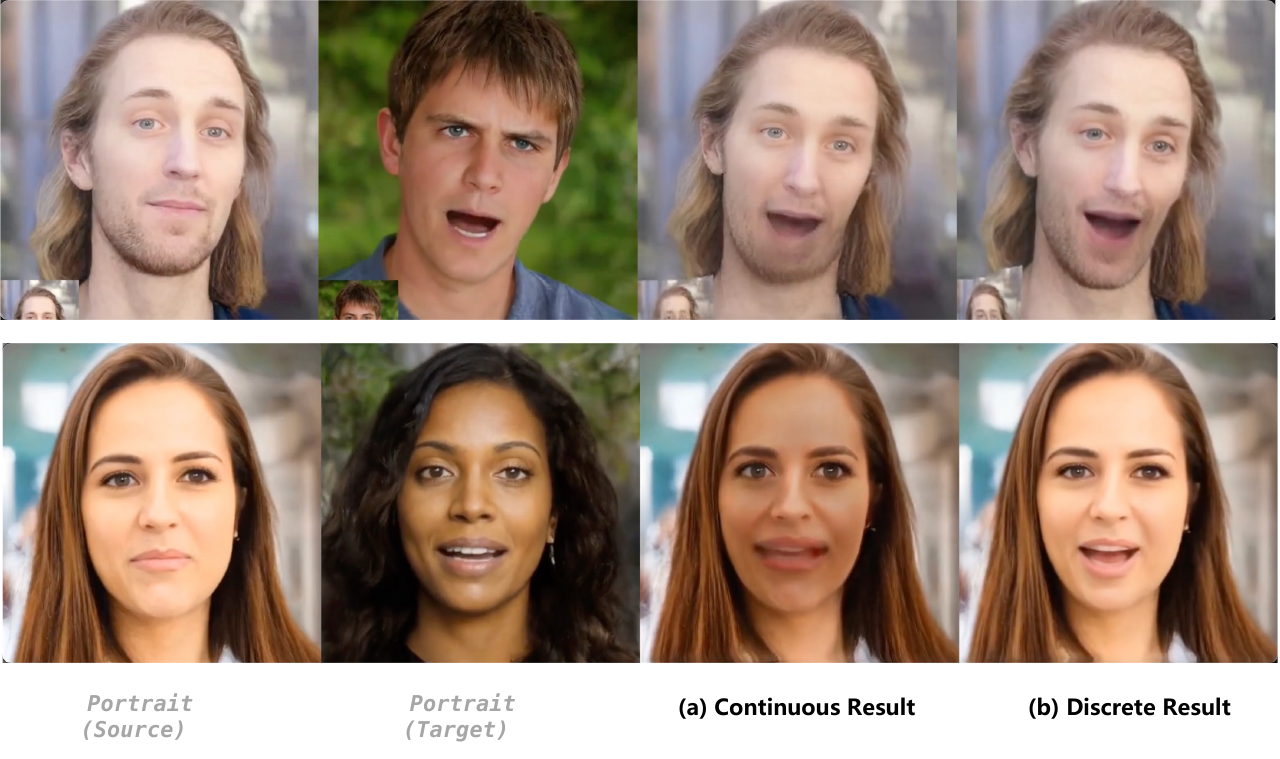}
\caption{Comparison of continuous (a) and discrete (b) representations in cross-identity facial animation. Continuous representation shows identity leakage, while discrete representation better preserves the target identity.}
\label{loss_function_ablation}
\end{figure}

\subsection{VQ as a Bottleneck Structure}
In contrast to recent studies~\cite{gaia, anitalker, vasa1, drobyshev2024emoportraits} that use continuous representations as bottleneck structures, we conducted a visualized ablation experiment to demonstrate that our discrete representation is less prone to information leakage in cross-identity driving scenarios compared to continuous representations. As shown in the figure, we replaced the discrete representation with a 20-dimensional representation and used photos of different identities to drive the single-frame image on the left. We can observe that the color and skin tone of the driving photos have `bled' into the driven photo, indicating that continuous representations suffer from identity leakage issues. Thanks to the efficient bottleneck design of VQ, this problem has been greatly alleviated.

\subsection{User Studies}

\begin{table}[htpb]

  \caption[Dataset statistics]{User Study on  Lip Sync (LS), Visual Appeal (VA), Prompt Following (PF), and Naturalness (N).}
  \label{tab:user_study}
  \centering
  \resizebox{1.0\linewidth}{!}{
  \begin{tabular}{@{}l cccc@{}} \toprule
  Method  & LS $\uparrow$  & VA $\uparrow$ & PF $\uparrow$ & N $\uparrow$ \\  \cmidrule{1-5} 

SadTalker~\cite{sadtalker}   & 2.25 &	 3.15 & 1.50 & 3.22  \\ 
EDTalker~\cite{edtalk}   & 3.58 &	 1.96 & 3.72 & 1.70  \\ 
AniTalker~\cite{anitalker}   & 2.56 &	 2.96 & 2.17 & 2.56  \\ 
EchoMimic~\cite{echomimic}   & 2.53 &	 \textbf{3.85} & 3.67 & 3.59  \\
\cmidrule{1-5} 
\textbf{VQTalker (Ours)}   & \textbf{4.08} &	 3.07 & \textbf{3.94} &\textbf{ 3.93}  \\ 

    \bottomrule
  \end{tabular}
}
\end{table}

We conducted a user study with 20 participants rating videos in different languages on a scale of 1 to 5 across four metrics: Lip Sync (LS), Visual Appeal (VA), Prompt Following (PF, including control of eye blinks, gaze, and head poses), and Naturalness (N). Results show that VQTalker achieves the highest scores in all categories except Visual Appeal, as shown in Table~\ref{tab:user_study}. Among the comparisons, EchoMimic~\cite{echomimic}, which is based on Stable Diffusion's redrawing technique, demonstrates superior visual quality.

\subsection{Limitations and Future Work}

While the VQTalker framework shows promising results in multilingual talking head generation, there are still areas for improvement and further exploration. For instance, in scenarios involving extreme facial movements, slight jitter may occasionally occur. This issue could potentially be mitigated by exploring higher-resolution quantization spaces or employing more advanced quantization techniques. Additionally, because we utilize a wrapping method, complex backgrounds or accessories can sometimes cause the background or the edges of the accessories to appear blurred. Future research could focus on addressing these challenges by refining the background handling and accessory integration methods.

\section{Ethical Consideration}

The rapid advancement of digital human technology, particularly in the creation of highly realistic virtual faces, presents significant ethical challenges. There are genuine concerns about the potential misuse of this technology for malicious purposes, such as deepfakes, identity theft, or the propagation of misinformation. To address these issues, it is crucial that developers and organizations establish comprehensive ethical guidelines before deploying such technologies. These guidelines should encompass principles of user privacy, data protection, and responsible use. Furthermore, to enhance accountability and prevent misuse, it is recommended to implement robust verification systems and content attribution methods for all digitally generated human representations. This could include blockchain-based authentication or secure metadata tagging. By proactively addressing these ethical considerations, we can foster the positive potential of digital human technology while minimizing its risks to individuals and society.

\section{Acknowledgments}

This work has been supported by the China NSFC Project (No. 92370206), Shanghai Municipal Science and Technology Major Project (2021SHZDZX0102) and the Key Research and Development Program of Jiangsu Province, China (Grant No. BE2022059).

\bibliography{aaai25}

\end{document}